\documentclass[
]{ceurart}

\usepackage{listings}
\usepackage{xcolor}
\usepackage{multicol}
\usepackage{multirow}
\usepackage{graphics}
\usepackage{hyperref}
\begin{document}

\copyrightyear{2022}
\copyrightclause{Copyright for this paper by its authors.
  Use permitted under Creative Commons License Attribution 4.0
  International (CC BY 4.0).}

\conference{Forum for Information Retrieval Evaluation, December 9-13, 2022, India}
\title{Indian Language Summarization using Pretrained Sequence-to-Sequence Models}

% \tnotetext[1]{You can use this document as the template for preparing your
%   publication. We recommend using the latest version of the ceurart style.
%   }

\author{Ashok Urlana}[%
email=ashok.urlana@research.iiit.ac.in,
]
% \cormark[1]
\address{Language Technologies Research Center, KCIS, IIIT Hyderabad, India}

\author{Sahil Manoj Bhatt}[%
email=sahil.bhatt@research.iiit.ac.in,
]
\author{Nirmal Surange}[%
email=nirmal.surange@research.iiit.ac.in,
]

\author{Manish Shrivastava}[%
email=m.shrivastava@iiit.ac.in,
]

\begin{abstract}
 The ILSUM shared task focuses on text summarization for two major Indian languages- Hindi and Gujarati, along with English. In this task, we experiment with various pretrained sequence-to-sequence models to find out the best model for each of the languages. We present a detailed overview of the models and our approaches in this paper. We secure the first rank across all three sub-tasks (English, Hindi and Gujarati). This paper also extensively analyzes the impact of k-fold cross-validation while experimenting with limited data size, and we also perform various experiments with a combination of the original and a filtered version of the data to determine the efficacy of the pretrained models. 
\end{abstract}

\begin{keywords}
    Indian language summarization \sep
    Sequence-to-Sequence models \sep
    Multilingual models \sep
\end{keywords}

\maketitle

\section{Introduction}
Automatic text summarization is a technique for obtaining a condensed version of a long document while retaining its relevance. The NLP community has become more interested in text summarization for Indian languages in recent years. The progress of text summarization has, however, been hindered due to the lack of high-quality datasets. Nevertheless, the availability of large-scale multilingual datasets such as XL-Sum\cite{hasan2021xl} and MassiveSumm\cite{varab-schluter-2021-massivesumm} have led to substantial progress in natural language generation and summarization tasks. Even though quality-wise, these datasets are far from perfect\cite{urlana-EtAl:2022:LREC}, they do serve as a good starting point in terms of quantity. Additionally, recent advancements in neural-based pretrained models have transformed the field significantly.

The goal of the ILSUM shared task is to create reusable corpora for Indian language summarization. The dataset is created by scraping the news articles and corresponding descriptions from publicly available news websites. ILSUM data\cite{fire2022-ilsum-acm, fire2022-ilsum} consists of a summarization corpus for two major Indian languages- Hindi and Gujarati, along with Indian English.  

This paper provides a comprehensive overview of the existing sequence-to-sequence models for Indian language and English summarization. For Hindi and Gujarati, we used multilingual models such as MT5\cite{xue2020mt5}, MBart\cite{liu2020multilingual} and IndicBART\cite{dabre2021indicbart} variants. We fine-tuned the PEGASUS\cite{zhang2020pegasus}, BART\cite{lewis2019bart}, T5\cite{raffel2020exploring} and ProphetNet\cite{qi2020prophetnet} models on English data. Out of all the models, for English, PEGASUS outperformed others, while for Hindi, MT5 gave us the best results, and for Gujarati, MBart performed the best. In order to avoid overfitting, we have performed k-fold cross-validation on the training dataset. We have observed that Hindi k-fold experiments had better scores than the experiments performed with the full version of the released data. We have applied several filters to assess the quality of the released datasets. Various combinations of filtered and original data were used to determine the efficacy of the pretrained generation models. We talk about our models, experiments and dataset filters later in this paper.

\section{Related Work}
Text summarization has been studied extensively, especially in the English language. Early research in summarization focused on extractive approaches, wherein summary sentences were chosen directly from the input text. On the other hand, abstractive approaches to summarization, such as neural attention models\cite{DBLP:journals/corr/RushCW15}, Seq2Seq RNNs \cite{DBLP:journals/corr/NallapatiXZ16}, Pointer-Generator networks \cite{DBLP:journals/corr/SeeLM17} focus on generating summaries that capture the meaning of the input text without necessarily choosing sentences directly from the text. With the emergence of large neural language models for generation tasks, abstractive approaches have become more popular and generate high-quality summaries. While there have been various improvements in model architectures and summarization techniques, a large part of the progress in English text summarization can be attributed to the availability of large-scale datasets, such as CNN/DailyMail\cite{DBLP:journals/corr/NallapatiXZ16, DBLP:journals/corr/HermannKGEKSB15}, Gigaword\cite{DBLP:journals/corr/RushCW15,graff2003english}, XSum\cite{Narayan2018DontGM}, etc.\\
This is in contrast to Indic languages, where little work has been done in summarization or related NLG tasks, such as headline generation. In recent times, however, there has been active research in this area, with the release of datasets such as XL-Sum\cite{hasan2021xl}, MassiveSumm\cite{varab-schluter-2021-massivesumm}, etc. These multilingual datasets consist of article-summary pairs from publicly available news domains, including Indian languages such as Hindi, Gujarati, Bengali, etc. The IndicNLG Suite\cite{kumar2022indicnlg} released datasets for several Indic language NLG tasks, such as sentence summarization and headline generation. More work needs to be done in this area to have models comparable to English summarization models in performance. 

\section{Corpus Description}
The dataset released for this task has been collected from several leading Indian newspaper websites. The English and Hindi datasets were scraped from indiatvnews\footnote{\url{https://www.indiatvnews.com/}}, and the Gujarati data was created by scraping the divyabhaskar\footnote{\url{https://www.divyabhaskar.co.in/}} and gujarati.news18\footnote{\url{https://gujarati.news18.com/}} websites. The Hindi and Gujarati datasets include articles/summaries which contain English words or phrases which have been code-mixed and script-mixed. Note that we have observed a few samples of English and Gujarati datasets, where the summaries consists of only one word. The ILSUM training data statistics are mentioned in Table~\ref{tab:data_stats}. We have used the Indic\cite{kunchukuttan2020indicnlp} tokenizer to generate the counts in Table~\ref{tab:data_stats}.
\begin{table*}[ht]
\caption{ILSUM Train Data Statistics}
\centering
\begin{tabular}{|c|cc|cc|cc|}
\hline
                & \multicolumn{2}{c|}{\textbf{English}} & \multicolumn{2}{c|}{\textbf{Hindi}} & \multicolumn{2}{c|}{\textbf{Gujarati}} \\ \hline
\#Pairs         & \multicolumn{2}{c|}{12564}            & \multicolumn{2}{c|}{7957}           & \multicolumn{2}{c|}{8457}              \\ \hline
                & \multicolumn{1}{c|}{Text}   & Summary & \multicolumn{1}{c|}{Text} & Summary & \multicolumn{1}{c|}{Text}    & Summary \\ \hline
\#Avg Words     & \multicolumn{1}{c|}{595}    & 36.24   & \multicolumn{1}{c|}{553}  & 40.17   & \multicolumn{1}{c|}{414.43}  & 32.26   \\ \hline
(Min, Max) Words     & \multicolumn{1}{c|}{(1, 5717)} & (1, 113) & \multicolumn{1}{c|}{(17, 5034)} & (6, 113) & \multicolumn{1}{c|}{(25, 2839)} & (1, 408) \\ \hline
\#Avg Sentences & \multicolumn{1}{c|}{10.29}  & 1.26    & \multicolumn{1}{c|}{18.1} & 1.7     & \multicolumn{1}{c|}{21.28}   & 1.57    \\ \hline
(Min, Max) Sentences & \multicolumn{1}{c|}{(1, 169)}  & (1, 17)  & \multicolumn{1}{c|}{(1, 157)}   & (1, 9)   & \multicolumn{1}{c|}{(1, 187)}   & (1, 46)  \\ \hline
\end{tabular}
\label{tab:data_stats}
\end{table*}

\section{Model Description}

The pretrained language models (PLMs) used for downstream tasks are pretrained using massive amounts of unlabeled text data. A PLM encodes extensive linguistic knowledge into a vast amount of parameters\cite{li2021pretrained}, which stimulates universal representations and improves generation quality. We have experimented with various pretrained generation models to find the optimal architecture. \\
\textbf{T5} \cite{raffel2020exploring} model proposes defining every NLP task in a text-to-text format. The model consists of an encoder-decoder Transformer architecture finetuned on the C4 corpus. In our experiments, we use both the T5-Base (220M parameters) and T5-Large (770M parameters) versions of the model. Since T5 is trained on an English-only dataset, we also look at the multilingual variants of the model for our experiments in Hindi and Gujarati. The MT5 model\cite{xue2020mt5} uses an architecture very similar to T5, and is trained on 101 languages, as described in the mC4 dataset. Owing to the large size of the models, we only finetuned the base version (580M parameters) of the MT5 model (the large version has 1.2B parameters).\\
\textbf{BART} \cite{lewis2019bart} is a denoising autoencoder for pretraining seq2seq models, which is similar to both BERT and GPT. Since it uses a bidirectional encoder like BERT, and an autoregressive decoder like GPT. The model was trained by corrupting the text using a noising function, and reconstructing the original text. We experiment with the BART-large model (406M parameters), and then also try out versions of the BART model finetuned on different datasets, namely the BART-Large-CNN and BART-Large-XSUM model, finetuned on the CNN-Daily Mail and XSUM datasets respectively. We try out multilingual variants\cite{liu2020multilingual} of the BART model for Hindi and Gujarati summarization experiments, namely the MBart-Large-50 (610M parameters) model\cite{DBLP:journals/corr/abs-2008-00401}, trained on 50 languages.\\
\textbf{PEGASUS} \cite{zhang2020pegasus} uses the extracted gap sentences (GSG) self-supervised objective strategy to train the encoder-decoder model. Rather than masking a smaller text span as in BART and T5, PEGASUS masks the entire sentence. Later, it concatenates the gap sentences into pseudo summaries. It chooses the sentences based on importance. In the same way as T5, PEGASUS does not reconstruct full sequence of inputs but only masked sentences. The pretraining is performed with C4\cite{raffel2020exploring} and HugeNews corpus. We finetune the PEGASUS-large model on the ILSUM English corpus.\\
\textbf{BRIO} \cite{liu2022brio} is a novel training paradigm to achieve neural abstractive summarization, wherein a contrastive learning component is introduced to reinforce the abstractive model’s ability to estimate the probability of system-generated summaries more precisely instead of using MLE training alone. Two stages are involved in this approach: the first stage generates the candidates using a pretrained sequence-to-sequence model, and next stage selects the best one.\\
\textbf{ProphetNet} \cite{qi2020prophetnet} introduced a novel self-supervised objective, wherein the goal is to predict the next-$n$ tokens, instead of just optimizing for one-step ahead predictions. We experiment with ProphetNet in our English summarization experiments.\\
\textbf{IndicBART}\cite{dabre2021indicbart} is a pretrained sequence-to-sequence model trained on 11 Indic languages and English. It follows the masked span reconstruction objective similar to MBart. In contrast to available generation models, IndicBART utilizes the orthographic similarity between the Indian languages to achieve better cross-lingual transfer learning capabilities. This model size (244M) is much smaller than MBart and MT5 models with compact vocabulary. We finetune the IndicBART model on Hindi and Gujarati datasets.\\
\textbf{Adapters:}
Recently proposed lightweight adapters\cite{pfeiffer2020adapterhub} are effective at mitigating the overhead of pretrained language models for downstream tasks. We can update the adapters during finetuning and freezing most of the PLM parameters. In recent work\cite{zhao2022adapt}, adapters were applied to perform Gujarati text summarization. Adapters can not only speed up training time but are also storage efficient since they require saving only adapter weights instead of entire finetuned model weights. 
\begin{table}[ht]
\caption{ILSUM Experiments on Validation Data. *Finetuned on the combination of Hindi and Gujarati Data}
\centering
\begin{tabular}{c|c|c|ccc}
\hline
 &
   &
  \textbf{} &
  \multicolumn{3}{c}{\textbf{Validation Scores}} \\ \hline
\textbf{Lang} &
  \textbf{Model} &
  \textbf{Full Data / k-fold} &
  \multicolumn{1}{c|}{\textbf{R-1}} &
  \multicolumn{1}{c|}{\textbf{R-2}} &
  \textbf{R-4} \\ \hline
\multirow{8}{*}{English} &
  PEGASUS &
  Full Data &
  \multicolumn{1}{c}{\textbf{56.85}} &
  \multicolumn{1}{c}{\textbf{45.92}} &
  \textbf{43.36} \\ 
 &
  T5$_{large}$ &
  Full Data &
  \multicolumn{1}{c}{56.05} &
  \multicolumn{1}{c}{45.03} &
  42.36 \\  
 &
  BART$_{large}$ &
  k-fold &
  \multicolumn{1}{c}{54.83} &
  \multicolumn{1}{c}{43.58} &
  40.71 \\  
 &
  PEGASUS xsum &
  Full Data &
  \multicolumn{1}{c}{54.66} &
  \multicolumn{1}{c}{43.48} &
  40.64 \\  
 &
  BRIO &
  Full Data &
  \multicolumn{1}{c}{53.57} &
  \multicolumn{1}{c}{41.86} &
  38.81 \\  
 &
  BART$_{large}$ xsum &
  k-fold &
  \multicolumn{1}{c}{53.35} &
  \multicolumn{1}{c}{41.74} &
  38.75 \\  
 &
  T5$_{base}$ + Adapter &
  k-fold &
  \multicolumn{1}{c}{51.91} &
  \multicolumn{1}{c}{40.07} &
  37.1 \\  
 &
  ProphetNet &
  k-fold &
  \multicolumn{1}{c}{49.51} &
  \multicolumn{1}{c}{36.98} &
  33.83 \\ \hline
\multirow{6}{*}{Hindi} &
  IndicBART &
  k-fold &
  \multicolumn{1}{c}{\textbf{60.73}} &
  \multicolumn{1}{c}{\textbf{51.26}} &
  \textbf{47.57} \\  
 &
  MT5$_{base}$ &
  k-fold &
  \multicolumn{1}{c}{60.04} &
  \multicolumn{1}{c}{50.72} &
  46.82 \\  
 &
  MT5$_{base}$* &
  Full Data &
  \multicolumn{1}{c}{58.65} &
  \multicolumn{1}{c}{49.09} &
  45.08 \\ 
 &
  IndicBART-SentSumm &
  k-fold &
  \multicolumn{1}{c}{58.09} &
  \multicolumn{1}{c}{47.99} &
  43.72 \\  
 &
  MBart$_{large}$50 + Adapters &
  Full Data &
  \multicolumn{1}{c}{56.26} &
  \multicolumn{1}{c}{45.56} &
  41.21 \\  
 &
  MBart$_{large}$50 &
  Full Data &
  \multicolumn{1}{c}{55.76} &
  \multicolumn{1}{c}{44.96} &
  40.59 \\ \hline
\multirow{6}{*}{Gujarati} &
  MBart$_{large}$50 &
  Full Data &
  \multicolumn{1}{c}{\textbf{26.20}} &
  \multicolumn{1}{c}{\textbf{16.44}} &
  \textbf{12.16} \\  
 &
  MT5$_{base}$ &
  Full Data &
  \multicolumn{1}{c}{25.11} &
  \multicolumn{1}{c}{15.81} &
  11.68 \\  
 &
%   MBart$_{Large}$CC25 &
%   Full Data &
%   \multicolumn{1}{c|}{-} &
%   \multicolumn{1}{c|}{-} &
%   - &
%   \multicolumn{1}{c|}{24.47} &
%   \multicolumn{1}{c|}{14.73} &
%   10.92 \\  
%  &
  MT5$_{base}$* &
  Full Data &
  \multicolumn{1}{c}{24.16} &
  \multicolumn{1}{c}{14.68} &
  10.79 \\  
 &
  IndicBART &
  k-fold &
  \multicolumn{1}{c}{23.38} &
  \multicolumn{1}{c}{13.34} &
  9.35 \\  
 &
  MBart$_{large}$50 + Adapter &
  Full Data &
  \multicolumn{1}{c}{21.63} &
  \multicolumn{1}{c}{13.04} &
  9.56 \\ \hline
\end{tabular}
\label{tab:val_scores}
\end{table}

\section{Experiments and Results}
We have performed experiments under two different settings: the first is with the entire released dataset (full data), and the other is where we split the dataset into 10 folds and utilize 90\% data (9 folds) for training and 10\% data (1 fold) for validation. In both settings, the released data in the validation phase was used for testing purposes and we report these results in Table~\ref{tab:val_scores}. Note that doing such k-fold cross validation experiments were also essential to evaluate our models' performance because validation summaries were not provided to us.\\
% We have finetuned several pretrained sequence-to-sequence models on ILSUM training data and obtained the validation scores to find the optimal architecture. Later, we experimented with only the best-performing models in the validation phase to obtain the test scores as mentioned in Table~\ref{tab:test_scores}. 
We use the standard ROUGE metric\cite{lin-2004-rouge} to compute all the scores. 
We observed that PEGASUS yields the best results for English when finetuned on the full data version in the validation phase. We achieved the best results when we finetuned IndicBART and MBart using k-fold and full data during the validation phase. Finetuning a model on k-fold data might sometimes lead to better results than finetuning it on the entire dataset, which indicates that the dataset needs to be studied more and appropriate filters need to be applied, to see which examples in the dataset contribute to the model learning something useful. We discuss this in the next section.\\
Based on the results of the validation phase, we submit results from the best models in the test phase. While PEGASUS and MBart still give us the best results for English and Gujarati respectively, MT5 performs better than IndicBART for Hindi when finetuned on k-fold data. Hyper-parameter settings are listed in Table~\ref{tab:parameters}. \\
The multilingual models have been pretrained on large amounts of data, and they are sufficiently capable of handling the presence of code-mixing in the dataset, which we observe in the outputs as well. The models generate good summaries and can add relevant English text in Hindi and Gujarati examples where appropriate. For instance, the average number of English words in Hindi and Gujarati training summaries is 0.25 and 1.91 respectively. For the test set released for Hindi and Gujarati, the summaries generated by our models have an average of 0.23 and 1.44 English words per summary. Note that the average number of English words in Hindi summaries is less because a large number of training samples are purely in Hindi and do not contain any English words or characters.

% \usepackage{multirow}
\begin{table}[ht]
\caption{ILSUM scores on Test Data}
\centering
\begin{tabular}{c|c|c|ccc}
\hline
                          &                & \textbf{} & \multicolumn{3}{c}{\textbf{Test Scores}}                                                 \\ \hline
\textbf{Lang} &
  \textbf{Model} &
  \textbf{Full Data / k-fold} &
  \multicolumn{1}{c|}{\textbf{R-1}} &
  \multicolumn{1}{c|}{\textbf{R-2}} &
  \textbf{R-4} \\ \hline
                          & PEGASUS        & Full Data & \multicolumn{1}{c|}{\textbf{55.83}} & \multicolumn{1}{c|}{\textbf{44.58}} & \textbf{41.8} \\  
\multirow{-2}{*}{English} & T5$_{large}$       & Full Data & \multicolumn{1}{c|}{54.73}          & \multicolumn{1}{c|}{43.08}          & 40.12         \\ \hline
 &
  MT5$_{base}$ &
  k-fold &
  \multicolumn{1}{c|}{\cellcolor[HTML]{FFFFFF}\textbf{60.72}} &
  \multicolumn{1}{c|}{\cellcolor[HTML]{FFFFFF}\textbf{51.02}} &
  \cellcolor[HTML]{FFFFFF}\textbf{47.11} \\ 
\multirow{-2}{*}{Hindi}   & IndicBART      & k-fold     & \multicolumn{1}{c|}{58.38}          & \multicolumn{1}{c|}{48.31}          & 44.25         \\ \hline
                          & MBart$_{large}$50 & Full Data & \multicolumn{1}{c|}{\textbf{26.11}} & \multicolumn{1}{c|}{16.51}          & 12.41         \\ 
\multirow{-2}{*}{Gujarati} &
  MBart$_{large}$50 &
  Full Data (dropout=0.2) &
  \multicolumn{1}{c|}{26.07} &
  \multicolumn{1}{c|}{\textbf{16.60}} &
  \textbf{12.58} \\ \hline
\end{tabular}
\label{tab:test_scores}
\end{table}

\begin{table}[ht]
\caption{Experimental setup and parameters settings}
\centering
\scalebox{0.9}{
\begin{tabular}{c|c|c|c|c|c|c|c|c}
\hline
\textbf{Parameters} & \textbf{BART} & \textbf{T5} & \textbf{ProphetNet} & \textbf{PEGASUS} & \textbf{BRIO} & \textbf{MBart} & \textbf{MT5} & \textbf{IndicBART} \\ \hline
Max source length & 512   & 512   & 512   & 512   & 512   & 512    & 512    & 512   \\ \hline
Max target length & 75    & 75    & 75    & 75    & 75    & 75     & 100    & 75    \\ \hline
Batch Size        & 2     & 1     & 1     & 2     & 2     & 4      & 2      & 2     \\ \hline
Epochs            & 5     & 5     & 5     & 5     & 5     & 5      & 10     & 10    \\ \hline
Vocab Size        & 50265 & 32128 & 30522 & 96103 & 50264 & 250054 & 250112 & 64015 \\ \hline
Beam Size         & 4     & 4     & 5     & 4     & 4     & 4      & 4      & 4     \\ \hline
Learning Rate       & 5e-5      & 5e-5    & 5e-5            & 5e-4         & 5e-5      & 5e-5       & 5e-5     & 5e-5           \\ \hline
\end{tabular}}
\label{tab:parameters}
\end{table}
\section{Data Quality Assessment}
To verify the quality of the data, we have applied some of the filters mentioned in TeSum\cite{urlana-EtAl:2022:LREC}.
Filters were applied include checking whether there are:
\begin{enumerate}
    \item Empty instances
    \item Duplicate pairs and summaries within the dataset
    \item Cases where the first few sentences of the article itself are taken as the summary
    \item Check whether the summary is `compressed enough', i.e., we should not have summaries comparable in size to the text that has to be summarized. Compression is a good measure of telling us if the summary provided is a shortened version of the input document/text or not.
\end{enumerate}

Filters counts for all the languages can be found in Table~\ref{tab:filtration}. It is important to note that, based on our filters, only about 68\% of the Hindi summaries are valid since many are simply the first few sentences of the article. It could also be one of the reasons for models giving better results on k-fold data. Some of the folds in the training data might contain a large percentage of high-quality, valid summaries while leaving out a significant number of summaries which we consider invalid. Note that for Gujarati and English, the number of final valid article-summary pairs is comparable to the original dataset size, which is why the top-performing models give better results when finetuned on the whole dataset as compared to k-fold subsets.
\begin{table}[ht]
\caption{Filtration counts of ILSUM data}
\centering
\begin{tabular}{c|c|c|c}
\hline
\textbf{Filters}           & \textbf{Hindi}   & \textbf{Gujarati} & \textbf{English} \\ \hline
\textbf{Dataset Size}               & 7957             & 8457              & 12565            \\ \hline
\textbf{Empty}                      & 0                & 0                 & 1                \\ \hline
\textbf{Duplicate Pairs}            & 23               & 0                 & 0                \\ \hline
\textbf{Duplicate Summary}          & 15               & 113               & 117              \\ \hline
\textbf{Prefixes}                   & 2518             & 135               & 486              \\ \hline
\textbf{Compression} \textless 50\% & 11               & 37                & 182              \\ \hline
\textbf{Final Valid}                & 5390             & 8172              & 11779            \\ \hline
\textbf{Valid \%}          & \textbf{67.74\%} & \textbf{96.63\%}  & \textbf{93.74\%} \\ \hline
\end{tabular}
\label{tab:filtration}
\end{table}

\begin{table}[h]
\centering
\caption{Validation set ROUGE scores on ILSUM corpus. This table reports the mean ROUGE scores and its standard deviation over 10 runs}
\label{tab:my-table}
\begin{tabular}{|c|c|c|c|c|c|}
\hline
\textbf{Lang}                       & \textbf{Model}                                           &    \textbf{Data composition}                      & \textbf{R-1} & \textbf{R-2} & \textbf{R-L} \\ \hline \hline
                                    &                                 & Original Data            & 52.51 ± 1.1  & 40.91 ± 1.36 & 47.81 ± 1.16 \\  
                                    &                                 & Original + Filtered Data & 51.65 ± 1.14 & 40.07 ± 1.25 & 46 ± 3.67    \\  
                                    &                                  & Filtered Data            & 51.88 ± 1.25 & 40.37 ± 1.39 & 47.32 ± 1.31 \\  
                                    & \multirow{-4}{*}{PEGASUS}        & Filtered + Original Data & 53.28 ± 1.18 & 41.82 ± 1.3  & 48.67 ± 1.2  \\ \cline{2-6} 
                                    &                                & Original Data            & \textbf{53.45 ± 0.95} & \textbf{42.16 ± 1.13} & \textbf{48.97 ± 1.05} \\ 
                                    &                                  & Original + Filtered Data & 53.22 ± 1.23 & 42.04 ± 1.41 & 48.85 ± 1.31 \\  
                                    &                                  & Filtered Data            & 51.9 ± 1.37  & 40.49 ± 1.53 & 47.38 ± 1.46 \\ 
                                    & \multirow{-4}{*}{T5-large}       & Filtered + Original Data & 53.33 ± 0.83 & 42.1 ± 0.96  & 48.92 ± 0.86 \\ \cline{2-6} 
                                    &                                  & Original Data            & 50.25 ± 1.52 & 38.15 ± 1.85 & 45.46 ± 1.63 \\ 
                                    &                                  & Original + Filtered Data & 51.42 ± 0.88 & 39.85 ± 1.11 & 46.93 ± 1    \\  
                                    &                                  & Filtered Data            & 51.21 ± 1.3  & 39.83 ± 1.57 & 46.79 ± 1.38 \\  
\multirow{-12}{*}{\textbf{English}} & \multirow{-4}{*}{BART-large}     & Filtered + Original Data & 52.45 ± 1.05 & 40.98 ± 1.29 & 48 ± 1.17    \\ \hline \hline
                                    &                                                          & Original Data            & 26.36 ± 1.02 & 12.66 ± 0.73 & 26.28 ± 0.98 \\ 
                                    &                                                          & Original + Filtered Data & 21.58 ± 0.66 & 9.84 ± 0.76  & 21.45 ± 0.6  \\  
                                    &                                                          & Filtered Data            & 21.27 ± 0.88 & 9.75 ± 0.56  & 21.12 ± 0.86 \\  
                                    & \multirow{-4}{*}{IndicBART}                              & Filtered + Original Data & 25.67 ± 1.04 & 12.16 ± 0.82 & 25.57 ± 1    \\ \cline{2-6} 
                                    &                                  & Original Data            & \textbf{27.04 ± 1.22} & \textbf{13.21 ± 0.61} & \textbf{26.96 ± 1.22} \\
                                    &                                  & Original + Filtered Data & 20.33 ± 0.91 & 9.26 ± 0.8   & 20.2 ± 0.92  \\  
                                    &                                  & Filtered Data            & 20.61 ± 1.55 & 9.47 ± 0.67  & 20.51 ± 1.53 \\  
\multirow{-8}{*}{\textbf{Hindi}}    & \multirow{-4}{*}{MT5-base}       & Filtered + Original Data & 26.73 ± 1.11 & 12.83 ± 0.61 & 26.64 ± 1.1  \\ \hline \hline
                                    &                                  & Original Data            & 20.36 ± 0.67 & 11.65 ± 1.13 & 20.01 ± 0.72 \\ 
                                    &                                  & Original + Filtered Data & 16.04 ± 1.12 & 9.23 ± 0.76  & 15.83 ± 1.15 \\ 
                                    &                                  & Filtered Data            & 12.82 ± 2.28 & 6.6 ± 1.54   & 12.38 ± 2.36 \\ 
                                    & \multirow{-4}{*}{MBart Large 50} & Filtered + Original Data & 19.55 ± 0.74 & 11.42 ± 0.43 & 19.2 ± 0.72  \\ \cline{2-6} 
                                    &                                  & Original Data            & \textbf{21.55 ± 0.77} & \textbf{11.81 ± 0.78} & \textbf{21.19 ± 0.83} \\  
                                    &                                  & Original + Filtered Data & 18.63 ± 0.93 & 9.23 ± 0.5   & 18.19 ± 0.92 \\  
                                    &                                  & Filtered Data            & 9.66 ± 0.97  & 4.84 ± 0.56  & 9.53 ± 0.92  \\ 
\multirow{-8}{*}{\textbf{Gujarati}} & \multirow{-4}{*}{MT5-base}       & Filtered + Original Data & 20.29 ± 0.62 & 10.7 ± 0.52  & 19.84 ± 0.56 \\ \hline
\end{tabular}
\label{tab:filter_exp}
\end{table}
\subsection{Data Variation Experiments}
The unavailability of large datasets is one of the main bottlenecks for neural models for text generation. The existing summarization datasets for Indian languages are quite small. To improve the model generation capabilities on limited dataset, we did k-fold cross-validation on the best performing models (see Table~\ref{tab:val_scores}). The mean ROUGE scores and standard deviation scores over 10 runs are reported in Table~\ref{tab:filter_exp}. We did 10-fold cross-validation using the released training dataset with the following combinations:
\begin{enumerate}
    \item \textbf{Original data:} Fine-tuned for 5 epochs with released training dataset
    \item \textbf{Original + Filtered data:} Finetuned for 3 epochs with original + 2 epochs with Filtered data
    \item \textbf{Filtered data:} Fine-tuned for 5 epochs with only filtered dataset
    \item \textbf{Filtered + Original data:} Finetuned for 3 epochs with filtered data + 2 epochs with original data
\end{enumerate}

To perform all the experiments, we used the `filtered data' obtained after applying filters mentioned in Table~\ref{tab:filtration}. To compare the models' performance on different variations of the training dataset, we have not made any changes in the validation data. As observed in Table~\ref{tab:filter_exp}, the experiments performed with `original' data produce better scores than the `filtered' data. Also, the models finetuned on the combination of the `filtered + original' dataset performed better compared to the `original+filtered' combination. 
\section{Discussion and Conclusions}
While having better models finetuned exclusively on Indian languages might benefit research in the area of Indian Language Summarization, creating larger, high-quality datasets for such languages will surely lead to progress in this field. It might be interesting to look at sources other than news websites as well, and to keep in mind the filters discussed earlier while creating the dataset.\\
For the ILSUM task, PEGASUS, MT5 and MBart give us the best results for English, Hindi and Gujarati respectively.
We conclude that the transformer-based pretrained seq2seq models are capable of generating high-quality summaries for the ILSUM shared task. 

\section*{Acknowledgements}
We thank the organizers of the ILSUM shared task for their help and support.

%% Define the bibliography file to be used
\bibliography{sample-ceur}

\begin{thebibliography}{26}
\expandafter\ifx\csname natexlab\endcsname\relax\def\natexlab#1{#1}\fi
\providecommand{\url}[1]{\texttt{#1}}
\providecommand{\href}[2]{#2}
\providecommand{\path}[1]{#1}
\providecommand{\DOIprefix}{doi:}
\providecommand{\ArXivprefix}{arXiv:}
\providecommand{\URLprefix}{URL: }
\providecommand{\Pubmedprefix}{pmid:}
\providecommand{\doi}[1]{\href{http://dx.doi.org/#1}{\path{#1}}}
\providecommand{\Pubmed}[1]{\href{pmid:#1}{\path{#1}}}
\providecommand{\bibinfo}[2]{#2}
\ifx\xfnm\relax \def\xfnm[#1]{\unskip,\space#1}\fi
%Type = Article
\bibitem[{Hasan et~al.(2021)Hasan, Bhattacharjee, Islam, Samin, Li, Kang,
  Rahman, and Shahriyar}]{hasan2021xl}
\bibinfo{author}{T.~Hasan}, \bibinfo{author}{A.~Bhattacharjee},
  \bibinfo{author}{M.~S. Islam}, \bibinfo{author}{K.~Samin},
  \bibinfo{author}{Y.-F. Li}, \bibinfo{author}{Y.-B. Kang},
  \bibinfo{author}{M.~S. Rahman}, \bibinfo{author}{R.~Shahriyar},
\newblock \bibinfo{title}{Xl-sum: Large-scale multilingual abstractive
  summarization for 44 languages},
\newblock \bibinfo{journal}{arXiv preprint arXiv:2106.13822}
  (\bibinfo{year}{2021}).
%Type = Inproceedings
\bibitem[{Varab and Schluter(2021)}]{varab-schluter-2021-massivesumm}
\bibinfo{author}{D.~Varab}, \bibinfo{author}{N.~Schluter},
\newblock \bibinfo{title}{{M}assive{S}umm: a very large-scale, very
  multilingual, news summarisation dataset},
\newblock in: \bibinfo{booktitle}{Proceedings of the 2021 Conference on
  Empirical Methods in Natural Language Processing},
  \bibinfo{publisher}{Association for Computational Linguistics},
  \bibinfo{address}{Online and Punta Cana, Dominican Republic},
  \bibinfo{year}{2021}, pp. \bibinfo{pages}{10150--10161}. \URLprefix
  \url{https://aclanthology.org/2021.emnlp-main.797}.
  \DOIprefix\doi{10.18653/v1/2021.emnlp-main.797}.
%Type = Inproceedings
\bibitem[{Urlana et~al.(2022)Urlana, Surange, Baswani, Ravva, and
  Shrivastava}]{urlana-EtAl:2022:LREC}
\bibinfo{author}{A.~Urlana}, \bibinfo{author}{N.~Surange},
  \bibinfo{author}{P.~Baswani}, \bibinfo{author}{P.~Ravva},
  \bibinfo{author}{M.~Shrivastava},
\newblock \bibinfo{title}{Tesum: Human-generated abstractive summarization
  corpus for telugu},
\newblock in: \bibinfo{booktitle}{Proceedings of the Language Resources and
  Evaluation Conference}, \bibinfo{publisher}{European Language Resources
  Association}, \bibinfo{address}{Marseille, France}, \bibinfo{year}{2022}, pp.
  \bibinfo{pages}{5712--5722}. \URLprefix
  \url{https://aclanthology.org/2022.lrec-1.614}.
%Type = Inproceedings
\bibitem[{Satapara et~al.(2022{\natexlab{a}})Satapara, Modha, Modha, and
  Mehta}]{fire2022-ilsum-acm}
\bibinfo{author}{S.~Satapara}, \bibinfo{author}{B.~Modha},
  \bibinfo{author}{S.~Modha}, \bibinfo{author}{P.~Mehta},
\newblock \bibinfo{title}{Fire 2022 ilsum track: Indian language
  summarization},
\newblock in: \bibinfo{booktitle}{Proceedings of the 14th Forum for Information
  Retrieval Evaluation}, \bibinfo{publisher}{{ACM}},
  \bibinfo{year}{2022}{\natexlab{a}}.
%Type = Inproceedings
\bibitem[{Satapara et~al.(2022{\natexlab{b}})Satapara, Modha, Modha, and
  Mehta}]{fire2022-ilsum}
\bibinfo{author}{S.~Satapara}, \bibinfo{author}{B.~Modha},
  \bibinfo{author}{S.~Modha}, \bibinfo{author}{P.~Mehta},
\newblock \bibinfo{title}{Findings of the first shared task on indian language
  summarization (ilsum): Approaches, challenges and the path ahead},
\newblock in: \bibinfo{booktitle}{Working Notes of {FIRE} 2022 - Forum for
  Information Retrieval Evaluation, Kolkata, India, December 9-13, 2022},
  {CEUR} Workshop Proceedings, \bibinfo{publisher}{CEUR-WS.org},
  \bibinfo{year}{2022}{\natexlab{b}}.
%Type = Article
\bibitem[{Xue et~al.(2020)Xue, Constant, Roberts, Kale, Al-Rfou, Siddhant,
  Barua, and Raffel}]{xue2020mt5}
\bibinfo{author}{L.~Xue}, \bibinfo{author}{N.~Constant},
  \bibinfo{author}{A.~Roberts}, \bibinfo{author}{M.~Kale},
  \bibinfo{author}{R.~Al-Rfou}, \bibinfo{author}{A.~Siddhant},
  \bibinfo{author}{A.~Barua}, \bibinfo{author}{C.~Raffel},
\newblock \bibinfo{title}{mt5: A massively multilingual pre-trained
  text-to-text transformer},
\newblock \bibinfo{journal}{arXiv preprint arXiv:2010.11934}
  (\bibinfo{year}{2020}).
%Type = Article
\bibitem[{Liu et~al.(2020)Liu, Gu, Goyal, Li, Edunov, Ghazvininejad, Lewis, and
  Zettlemoyer}]{liu2020multilingual}
\bibinfo{author}{Y.~Liu}, \bibinfo{author}{J.~Gu}, \bibinfo{author}{N.~Goyal},
  \bibinfo{author}{X.~Li}, \bibinfo{author}{S.~Edunov},
  \bibinfo{author}{M.~Ghazvininejad}, \bibinfo{author}{M.~Lewis},
  \bibinfo{author}{L.~Zettlemoyer},
\newblock \bibinfo{title}{Multilingual denoising pre-training for neural
  machine translation},
\newblock \bibinfo{journal}{Transactions of the Association for Computational
  Linguistics} \bibinfo{volume}{8} (\bibinfo{year}{2020})
  \bibinfo{pages}{726--742}.
%Type = Article
\bibitem[{Dabre et~al.(2021)Dabre, Shrotriya, Kunchukuttan, Puduppully, Khapra,
  and Kumar}]{dabre2021indicbart}
\bibinfo{author}{R.~Dabre}, \bibinfo{author}{H.~Shrotriya},
  \bibinfo{author}{A.~Kunchukuttan}, \bibinfo{author}{R.~Puduppully},
  \bibinfo{author}{M.~M. Khapra}, \bibinfo{author}{P.~Kumar},
\newblock \bibinfo{title}{Indicbart: A pre-trained model for natural language
  generation of indic languages},
\newblock \bibinfo{journal}{arXiv preprint arXiv:2109.02903}
  (\bibinfo{year}{2021}).
%Type = Inproceedings
\bibitem[{Zhang et~al.(2020)Zhang, Zhao, Saleh, and Liu}]{zhang2020pegasus}
\bibinfo{author}{J.~Zhang}, \bibinfo{author}{Y.~Zhao},
  \bibinfo{author}{M.~Saleh}, \bibinfo{author}{P.~Liu},
\newblock \bibinfo{title}{Pegasus: Pre-training with extracted gap-sentences
  for abstractive summarization},
\newblock in: \bibinfo{booktitle}{International Conference on Machine
  Learning}, \bibinfo{organization}{PMLR}, \bibinfo{year}{2020}, pp.
  \bibinfo{pages}{11328--11339}.
%Type = Article
\bibitem[{Lewis et~al.(2019)Lewis, Liu, Goyal, Ghazvininejad, Mohamed, Levy,
  Stoyanov, and Zettlemoyer}]{lewis2019bart}
\bibinfo{author}{M.~Lewis}, \bibinfo{author}{Y.~Liu},
  \bibinfo{author}{N.~Goyal}, \bibinfo{author}{M.~Ghazvininejad},
  \bibinfo{author}{A.~Mohamed}, \bibinfo{author}{O.~Levy},
  \bibinfo{author}{V.~Stoyanov}, \bibinfo{author}{L.~Zettlemoyer},
\newblock \bibinfo{title}{Bart: Denoising sequence-to-sequence pre-training for
  natural language generation, translation, and comprehension},
\newblock \bibinfo{journal}{arXiv preprint arXiv:1910.13461}
  (\bibinfo{year}{2019}).
%Type = Article
\bibitem[{Raffel et~al.(2020)Raffel, Shazeer, Roberts, Lee, Narang, Matena,
  Zhou, Li, Liu et~al.}]{raffel2020exploring}
\bibinfo{author}{C.~Raffel}, \bibinfo{author}{N.~Shazeer},
  \bibinfo{author}{A.~Roberts}, \bibinfo{author}{K.~Lee},
  \bibinfo{author}{S.~Narang}, \bibinfo{author}{M.~Matena},
  \bibinfo{author}{Y.~Zhou}, \bibinfo{author}{W.~Li}, \bibinfo{author}{P.~J.
  Liu}, et~al.,
\newblock \bibinfo{title}{Exploring the limits of transfer learning with a
  unified text-to-text transformer.},
\newblock \bibinfo{journal}{J. Mach. Learn. Res.} \bibinfo{volume}{21}
  (\bibinfo{year}{2020}) \bibinfo{pages}{1--67}.
%Type = Article
\bibitem[{Qi et~al.(2020)Qi, Yan, Gong, Liu, Duan, Chen, Zhang, and
  Zhou}]{qi2020prophetnet}
\bibinfo{author}{W.~Qi}, \bibinfo{author}{Y.~Yan}, \bibinfo{author}{Y.~Gong},
  \bibinfo{author}{D.~Liu}, \bibinfo{author}{N.~Duan},
  \bibinfo{author}{J.~Chen}, \bibinfo{author}{R.~Zhang},
  \bibinfo{author}{M.~Zhou},
\newblock \bibinfo{title}{Prophetnet: Predicting future n-gram for
  sequence-to-sequence pre-training},
\newblock \bibinfo{journal}{arXiv preprint arXiv:2001.04063}
  (\bibinfo{year}{2020}).
%Type = Article
\bibitem[{Rush et~al.(2015)Rush, Chopra, and
  Weston}]{DBLP:journals/corr/RushCW15}
\bibinfo{author}{A.~M. Rush}, \bibinfo{author}{S.~Chopra},
  \bibinfo{author}{J.~Weston},
\newblock \bibinfo{title}{A neural attention model for abstractive sentence
  summarization},
\newblock \bibinfo{journal}{CoRR} \bibinfo{volume}{abs/1509.00685}
  (\bibinfo{year}{2015}). \URLprefix \url{http://arxiv.org/abs/1509.00685}.
  \href{http://arxiv.org/abs/1509.00685}{{\tt arXiv:1509.00685}}.
%Type = Article
\bibitem[{Nallapati et~al.(2016)Nallapati, Xiang, and
  Zhou}]{DBLP:journals/corr/NallapatiXZ16}
\bibinfo{author}{R.~Nallapati}, \bibinfo{author}{B.~Xiang},
  \bibinfo{author}{B.~Zhou},
\newblock \bibinfo{title}{Sequence-to-sequence rnns for text summarization},
\newblock \bibinfo{journal}{CoRR} \bibinfo{volume}{abs/1602.06023}
  (\bibinfo{year}{2016}). \URLprefix \url{http://arxiv.org/abs/1602.06023}.
  \href{http://arxiv.org/abs/1602.06023}{{\tt arXiv:1602.06023}}.
%Type = Article
\bibitem[{See et~al.(2017)See, Liu, and Manning}]{DBLP:journals/corr/SeeLM17}
\bibinfo{author}{A.~See}, \bibinfo{author}{P.~J. Liu}, \bibinfo{author}{C.~D.
  Manning},
\newblock \bibinfo{title}{Get to the point: Summarization with
  pointer-generator networks},
\newblock \bibinfo{journal}{CoRR} \bibinfo{volume}{abs/1704.04368}
  (\bibinfo{year}{2017}). \URLprefix \url{http://arxiv.org/abs/1704.04368}.
  \href{http://arxiv.org/abs/1704.04368}{{\tt arXiv:1704.04368}}.
%Type = Article
\bibitem[{Hermann et~al.(2015)Hermann, Kocisk{\'{y}}, Grefenstette, Espeholt,
  Kay, Suleyman, and Blunsom}]{DBLP:journals/corr/HermannKGEKSB15}
\bibinfo{author}{K.~M. Hermann}, \bibinfo{author}{T.~Kocisk{\'{y}}},
  \bibinfo{author}{E.~Grefenstette}, \bibinfo{author}{L.~Espeholt},
  \bibinfo{author}{W.~Kay}, \bibinfo{author}{M.~Suleyman},
  \bibinfo{author}{P.~Blunsom},
\newblock \bibinfo{title}{Teaching machines to read and comprehend},
\newblock \bibinfo{journal}{CoRR} \bibinfo{volume}{abs/1506.03340}
  (\bibinfo{year}{2015}). \URLprefix \url{http://arxiv.org/abs/1506.03340}.
  \href{http://arxiv.org/abs/1506.03340}{{\tt arXiv:1506.03340}}.
%Type = Article
\bibitem[{Graff et~al.(2003)Graff, Kong, Chen, and Maeda}]{graff2003english}
\bibinfo{author}{D.~Graff}, \bibinfo{author}{J.~Kong},
  \bibinfo{author}{K.~Chen}, \bibinfo{author}{K.~Maeda},
\newblock \bibinfo{title}{English gigaword},
\newblock \bibinfo{journal}{Linguistic Data Consortium, Philadelphia}
  \bibinfo{volume}{4} (\bibinfo{year}{2003}) \bibinfo{pages}{34}.
%Type = Article
\bibitem[{Narayan et~al.(2018)Narayan, Cohen, and Lapata}]{Narayan2018DontGM}
\bibinfo{author}{S.~Narayan}, \bibinfo{author}{S.~B. Cohen},
  \bibinfo{author}{M.~Lapata},
\newblock \bibinfo{title}{Don't give me the details, just the summary!
  topic-aware convolutional neural networks for extreme summarization},
\newblock \bibinfo{journal}{ArXiv} \bibinfo{volume}{abs/1808.08745}
  (\bibinfo{year}{2018}).
%Type = Misc
\bibitem[{Kumar et~al.(2022)Kumar, Shrotriya, Sahu, Dabre, Puduppully,
  Kunchukuttan, Mishra, Khapra, and Kumar}]{kumar2022indicnlg}
\bibinfo{author}{A.~Kumar}, \bibinfo{author}{H.~Shrotriya},
  \bibinfo{author}{P.~Sahu}, \bibinfo{author}{R.~Dabre},
  \bibinfo{author}{R.~Puduppully}, \bibinfo{author}{A.~Kunchukuttan},
  \bibinfo{author}{A.~Mishra}, \bibinfo{author}{M.~M. Khapra},
  \bibinfo{author}{P.~Kumar}, \bibinfo{title}{Indicnlg suite: Multilingual
  datasets for diverse nlg tasks in indic languages}, \bibinfo{year}{2022}.
  \href{http://arxiv.org/abs/2203.05437}{{\tt arXiv:2203.05437}}.
%Type = Misc
\bibitem[{Kunchukuttan(2020)}]{kunchukuttan2020indicnlp}
\bibinfo{author}{A.~Kunchukuttan}, \bibinfo{title}{{The IndicNLP Library}},
  \bibinfo{howpublished}{\url{https://github.com/anoopkunchukuttan/indic_nlp_library/blob/master/docs/indicnlp.pdf}},
  \bibinfo{year}{2020}.
%Type = Article
\bibitem[{Li et~al.(2021)Li, Tang, Zhao, and Wen}]{li2021pretrained}
\bibinfo{author}{J.~Li}, \bibinfo{author}{T.~Tang}, \bibinfo{author}{W.~X.
  Zhao}, \bibinfo{author}{J.-R. Wen},
\newblock \bibinfo{title}{Pretrained language models for text generation: A
  survey},
\newblock \bibinfo{journal}{arXiv preprint arXiv:2105.10311}
  (\bibinfo{year}{2021}).
%Type = Article
\bibitem[{Tang et~al.(2020)Tang, Tran, Li, Chen, Goyal, Chaudhary, Gu, and
  Fan}]{DBLP:journals/corr/abs-2008-00401}
\bibinfo{author}{Y.~Tang}, \bibinfo{author}{C.~Tran}, \bibinfo{author}{X.~Li},
  \bibinfo{author}{P.~Chen}, \bibinfo{author}{N.~Goyal},
  \bibinfo{author}{V.~Chaudhary}, \bibinfo{author}{J.~Gu},
  \bibinfo{author}{A.~Fan},
\newblock \bibinfo{title}{Multilingual translation with extensible multilingual
  pretraining and finetuning},
\newblock \bibinfo{journal}{CoRR} \bibinfo{volume}{abs/2008.00401}
  (\bibinfo{year}{2020}). \URLprefix \url{https://arxiv.org/abs/2008.00401}.
  \href{http://arxiv.org/abs/2008.00401}{{\tt arXiv:2008.00401}}.
%Type = Article
\bibitem[{Liu et~al.(2022)Liu, Liu, Radev, and Neubig}]{liu2022brio}
\bibinfo{author}{Y.~Liu}, \bibinfo{author}{P.~Liu}, \bibinfo{author}{D.~Radev},
  \bibinfo{author}{G.~Neubig},
\newblock \bibinfo{title}{Brio: Bringing order to abstractive summarization},
\newblock \bibinfo{journal}{arXiv preprint arXiv:2203.16804}
  (\bibinfo{year}{2022}).
%Type = Article
\bibitem[{Pfeiffer et~al.(2020)Pfeiffer, R{\"u}ckl{\'e}, Poth, Kamath,
  Vuli{\'c}, Ruder, Cho, and Gurevych}]{pfeiffer2020adapterhub}
\bibinfo{author}{J.~Pfeiffer}, \bibinfo{author}{A.~R{\"u}ckl{\'e}},
  \bibinfo{author}{C.~Poth}, \bibinfo{author}{A.~Kamath},
  \bibinfo{author}{I.~Vuli{\'c}}, \bibinfo{author}{S.~Ruder},
  \bibinfo{author}{K.~Cho}, \bibinfo{author}{I.~Gurevych},
\newblock \bibinfo{title}{Adapterhub: A framework for adapting transformers},
\newblock \bibinfo{journal}{arXiv preprint arXiv:2007.07779}
  (\bibinfo{year}{2020}).
%Type = Article
\bibitem[{Zhao and Chen(2022)}]{zhao2022adapt}
\bibinfo{author}{Z.~Zhao}, \bibinfo{author}{P.~Chen},
\newblock \bibinfo{title}{To adapt or to fine-tune: A case study on abstractive
  summarization},
\newblock \bibinfo{journal}{arXiv preprint arXiv:2208.14559}
  (\bibinfo{year}{2022}).
%Type = Inproceedings
\bibitem[{Lin(2004)}]{lin-2004-rouge}
\bibinfo{author}{C.-Y. Lin},
\newblock \bibinfo{title}{{ROUGE}: A package for automatic evaluation of
  summaries},
\newblock in: \bibinfo{booktitle}{Text Summarization Branches Out},
  \bibinfo{publisher}{Association for Computational Linguistics},
  \bibinfo{address}{Barcelona, Spain}, \bibinfo{year}{2004}, pp.
  \bibinfo{pages}{74--81}. \URLprefix \url{https://aclanthology.org/W04-1013}.

\end{thebibliography}

% \appendix


% The sources for the ceur-art style are available via
% \begin{itemize}
% \item \href{https://github.com/yamadharma/ceurart}{GitHub},
% % \item \href{https://www.overleaf.com/project/5e76702c4acae70001d3bc87}{Overleaf},
% \item
%   \href{https://www.overleaf.com/latex/templates/template-for-submissions-to-ceur-workshop-proceedings-ceur-ws-dot-org/pkfscdkgkhcq}{Overleaf
%     template}.
% \end{itemize}

\end{document}